\documentclass[sigconf,natbib=false,nonacm]{acmart}
\AtBeginDocument{%
  }

\makeatletter\renewcommand\paragraph{\@startsection{paragraph}{4}{\z@}
  {.5em \@plus1ex \@minus.2ex}{-.5em}{\normalfont\normalsize\bfseries}}\makeatother

\RequirePackage[
    backend=biber,
  datamodel=acmdatamodel,
  style=acmnumeric,
  ]{biblatex}

\usepackage{amssymb}

\newcommand{\R}{\mathbb{R}}

\usepackage{amsmath}

\DeclareUnicodeCharacter{0301}{{~~}}

\definecolor{pink}{rgb}{1.0, 0.75, 0.8}
\definecolor{piggypink}{rgb}{0.99, 0.87, 0.9}
\usepackage{soul} 
\sethlcolor{pink}

\makeatletter
\def\@copyrightspace{\relax}
\makeatother

\begin{document}

\title{What can we Learn by Predicting Accuracy?}

\author{Olivier RISSER-MAROIX}
\authornote{Both authors contributed equally to this research.}
\email{orissermaroix@gmail.com}
\affiliation{
  \institution{LIPADE, Université Paris Cité} 
  \country{France}}

\author{Benjamin CHAMAND}
\authornotemark[1]
\email{benjamin.chamand@irit.fr} 
\affiliation{
  \institution{IRIT, Université de Toulouse, CNRS, Toulouse INP, UT3} 
  \country{France} 
}

\renewcommand{\shortauthors}{O. Risser-Maroix, B. Chamand}

\begin{abstract}
\noindent
    This paper seeks to answer the following question: \textit{"What can we learn by predicting accuracy?"}. 
    Indeed, classification is one of the most popular tasks in machine learning, and many loss functions have been developed to maximize this non-differentiable objective function.
    Unlike past work on loss function design, which was guided mainly by intuition and theory before being validated by experimentation, here we propose to approach this problem in the opposite way: we seek to extract knowledge by experimentation. 
    This data-driven approach is similar to that used in physics to discover general laws from data.
    We used a symbolic regression method to automatically find a mathematical expression highly correlated with a linear classifier's accuracy.
    The formula discovered on more than 260 datasets of embeddings has a Pearson's correlation of 0.96 and a $r^2$ of 0.93.
    More interestingly, this formula is highly explainable and confirms insights from various previous papers on loss design.
    We hope this work will open new perspectives in the search for new heuristics leading to a deeper understanding of machine learning theory. 
\end{abstract}



\maketitle

\section{Introduction}

Most work in machine learning is done by building up and evaluating components from theoretical intuitions.
Here we propose a different approach, which is to acquire insights from experimentation, in the same way that physicists have attempted to discover the analytical laws underlying physical phenomena in nature from observations. 
However, thanks to breakthroughs 
in artificial intelligence, a new trend to automate and assist research with \textit{Machine Learning} (ML) tools is emerging.
Some researchers started to use it in mathematics \cite{davies2021advancing} and physics \cite{douglas2022machine, schmidt2009distilling}.

In ML, the most similar setting would be the \textit{meta-learning} one.
In this \textit{learning-to-learn} paradigm, a model gains experience over multiple learning episodes and uses this experience to improve its future learning performance.
 Hospedales et al. \cite{hospedales2020meta} reported successful applications of meta-learning on diverse tasks such as hyperparameter optimization, neural architecture search (NAS), etc.
In this setting, the machine generally improves solutions without any human intervention.
Although meta-learning has been widely explored and is actively involved in increasing the performance of machine learning models. The solutions found are generally non-interpretable. 
So surprisingly, including AI in the process hasn't caught the attention as a tool helping in the theoretical discoveries of ML studies.
Hence, we investigate how machine learning can be integrated into the research process and lead us to better understand our discipline. 
As example, we propose here to tackle the problem of finding the key components of embeddings leading to better accuracy. 
This task could help us to better understand the intrinsic mechanisms of learning representations. 
Indeed, representation learning is often evaluated on benchmarks, such as \cite{hu2020xtreme} in NLP  or \cite{goyal2019scaling, zhai2019visual} in Computer Vision, where the task of classification is highly present. 
For example, self-supervised learning image representations are evaluated with a linear classifier.
Classification performances are generally measured using the \textit{accuracy}. 
To optimize this non-differentiable objective, researchers proposed proxy losses such as cross-entropy, hinge loss, and variants satisfying some properties and correcting several defaults of the previous ones. 
We can thus benefit from decades of research to validate the machine-generated function.

The task of predicting the future accuracy of a machine learning model has received little attention.
While this question may look odd at first glance, answering it has multiple applications, such as: fastening NAS by being able to predict the performance of a random architecture without having to train it \cite{istrate2019tapas, wen2020neural}; evaluating the accuracy of a classifier on an unlabeled test set \cite{deng2021labels}; or measuring the difficulty of a dataset \cite{Collins2018EvolutionaryDM, scheidegger2021efficient}.
Accuracy can thus be estimated from network weights \cite{yamada2016weight}, network architecture \cite{wen2020neural} or, as in our case, dataset statistics \cite{bensusan2001estimating, Collins2018EvolutionaryDM, deng2021labels}. 
Previous works mostly rely on regression models such as neural networks or random forests, making solutions found non-explainable \cite{deng2021labels, yamada2016weight}. 
While showing good performance for their respective use cases, those works did not focus on the interpretability of their solution.

In this paper, we provide a general formula by studying more than 260 datasets of embeddings with very different characteristics (size, dimension, number of classes, etc.). 
We propose to project those datasets into the same representation space by describing them as a set of statistics.
From those statistical representations, we found a formula able to predict the future classification performance of a linear classifier with a strong Pearson's correlation and $r^2$ score. 
When comparing similar pipelines, we found our formula simpler and more explainable. 
Finally, we analyze it in light of decades of research.

\section{Related Works}

The scientific method requires understanding the mathematical relationships between variables in a given system. 
Symbolic Regression (SR) aims to find a function that explains hidden relationships in data without having any prior knowledge of the function's form. 
On the other hand, traditional regression imposes a single fixed model structure during training, frequently chosen to be expressive (e.g., neural network, random forest, etc.) at the expense of being easily interpretable. 
Because SR is believed to be an NP-hard problem \cite{virgolin2022symbolic}, evolutionary methods have been developed to obtain approximate solutions \cite{Koza1994, koza1997genetic, augusto2000symbolic, lu2016using}.
The symbolic regression challenge has recently regained popularity, and novel approaches combining classical genetic programming and modern deep reinforcement learning have emerged \cite{petersen2021deep, landajuela2021discovering, mundhenk2021seeding, udrescu2020ai, virgolin2021improving}. 
Indeed, when tested on 240 small datasets of 250 observations, SR was found to be both highly interpretable and competitive on small datasets \cite{Wilstrup2021SymbolicRO}.

To learn a model mapping any dataset to a predicted accuracy score, we must build a shared representation space among all datasets. 
For example, \cite{mansilla2004classifier} used nine metrics of data complexity to characterize the behavior of several classifiers (linear, KNN, etc.) and thus found their respective domains of competence: where they perform best.
In another work \cite{ho2002complexity} found, by analyzing the twelve measures they proposed, that rich structures exist in such a measurement space,  revealing the intricate relationship among the factors affecting the difficulty of a problem.
However, they only examined the training sets' structures without generalizing to unseen points. 
More recently, \cite{lorena2019complex} listed 22 complexity measures from past literature.
Unfortunately, most have a complexity greater than $O(n^2)$, with $n$ the number of points in the dataset. 
This makes those measures difficult to scale up to larger datasets. 
The task of finding statistical features for datasets representation is still considered an open question \cite{deng2021labels}.

Close to our work, \cite{Collins2018EvolutionaryDM} propose to understand the difficulty of a text classification task using textual statistics to describe datasets used and a genetic algorithm to find the summation of those statistics correlating best with the F1-score. 
However, their work is limited by the choice of features, such as $n$-grams, making it only usable for textual datasets. 
By searching, with a \textit{Genetic Algorithm} (GA),  an unweighted summation of a subset of proposed statistics, they could only cover features having the same magnitude, discarding pertinent other ones such as the dataset size. 
The choice of an unweighted summation is likely to perform worse than a weighted one learned by a linear regression model. 
However, our solution confirmed intuition from \cite{ho2002complexity} suggesting that the relationship between statistics is highly non-linear.

In another interesting work, \cite{bensusan2001estimating} proposed estimating the predictive accuracy of several classifiers to select the most suited for a given dataset.
In their analysis, the authors studied only one linear model: the linear discriminant analysis (LDA).
However, the current state-of-the-art use \textit{softmax} based models require gradient descent approaches. 
In order to extract knowledge from a meta-dataset of tabular datasets, they used Cubist\footnote{\url{https://cran.r-project.org/web/packages/Cubist/vignettes/cubist.html}}, a package producing models in the form of rulesets. 
However, by being numerous and formulated with hard-coded values, the generated rules are complex and difficult to generalize.

By being applied to specific data such as text or tabular ones, neither \cite{Collins2018EvolutionaryDM} nor \cite{bensusan2001estimating} used the same set of statistics, making the results of their proposed pipeline not comparable. 
Here, we focus on general embeddings from datasets with a broader diversity in their characteristics, such as the range of the number of classes (\cite{Collins2018EvolutionaryDM} the biggest one being 115 while we generalize up to 1824 classes). 
In this work, we choose to describe our datasets with 19 domain agnostic statistics. 
Moreover, we compare the solution found by our pipeline with solutions found with previous ones \cite{Collins2018EvolutionaryDM, bensusan2001estimating}. 
Using our set of general statistics, we found that our solution is more efficient than the others while being simpler.

\begin{figure}
\begin{center}
\centering
    \includegraphics[width=.47\textwidth]{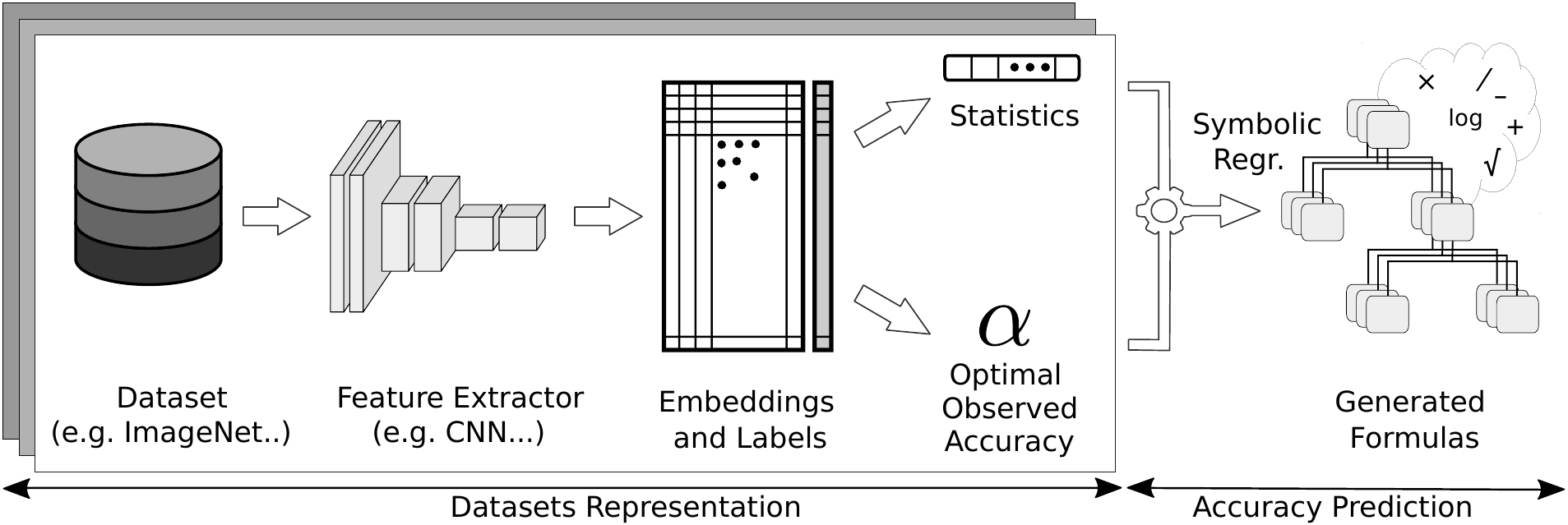}
    \caption{Proposed workflow}
    \label{fig:workflow_gray}
\end{center}
\end{figure}

\section{Proposed Approach}

As illustrated in Figure~\ref{fig:workflow_gray}, our method is composed of two parts: (1) the creation of a meta-dataset $\mathcal{M}$ from the combination of different datasets and feature extractors, its representation, and ground-truth creation; (2) the discovery of an explainable heuristic by symbolic regression modeling.
We detail each component in the following paragraphs.

\paragraph{Datasets and Feature Extractors} 
To find a {general law} covering a wide range of cases for a classification task, we selected 12 datasets and 22 feature extractors. 
The number of classes ranges from 10 to 1854, while the dimensionality of the features ranges from 256 to 2048. 
The selected datasets are MNIST \cite{lecun2010mnist}, 
CIFAR10 \cite{Krizhevsky09learningmultiple}, 
DTD \cite{cimpoi14describing}, 
PhotoArt \cite{wu2014learning}, 
CIFAR100 \cite{Krizhevsky09learningmultiple}, 
105-PinterestFaces \cite{kagglePinsFace}, 
CUB200 \cite{WelinderEtal2010}, 
ImageNet-R \cite{hendrycks2020many}, 
Caltech256 \cite{caltech256}, 
FSS1000 \cite{FSS1000}, 
ImageNetMini \cite{Le2015TinyIV},
THINGS \cite{Hebart2019THINGS}, 
containing respectively 10, 10, 47, 50, 100, 105, 200, 200, 256, 1000, 1000, 1854 classes.
Regarding the feature extractors, different architectures have been selected with different pretraining to cover a large number of dimensions and difficulty levels of linear classification.
For example, an architecture like FaceNet \cite{schroff2015facenet} is expected to perform poorly on CIFAR datasets since it is learned on a face recognition task while being a better feature extractor on this same dataset than a random initialized one. 
The ImageNet pretrained feature extractors used are: AlexNet \cite{krizhevsky2012imagenet},  ResNet \cite{he2016deep} (RN-\{18, 50, 101\}), DenseNet \cite{huang2017densely} (DN-\{169, 201\}), SqueezeNet \cite{iandola2016squeezenet}, MobileNetv2 \cite{sandler2018mobilenetv2}, MobileNetv3 \cite{howard2019searching} small and large versions. 
We also used FaceNet \cite{schroff2015facenet} pretrained on VGGFaces2 
and CLIP-\{RN50, ViT16b, ViT32b\} \cite{radford2021learning} pretrained on millions of image-text pairs.
As untrained feature extractors, we used: 
ResNet (RN-\{34, 152\}), 
DenseNet (DN-\{169, 201\}), 
SqueezeNet,  
MobileNetv2,  
MobileNetv3 
small and large versions.
All embedding dimensions represented here are: \{256, 512, 576, 768, 960, 1024, 1280, 1664, 1792, 1920, 2048\}. 
We refer to embeddings produced from the combination of all datasets of images by all feature extractors as a \textit{dataset of embeddings}. 
We construct a meta-dataset $\mathcal{M}$ from those 260+ datasets of embeddings.

\begin{figure}
    \centering
    \includegraphics[width=.47\textwidth]{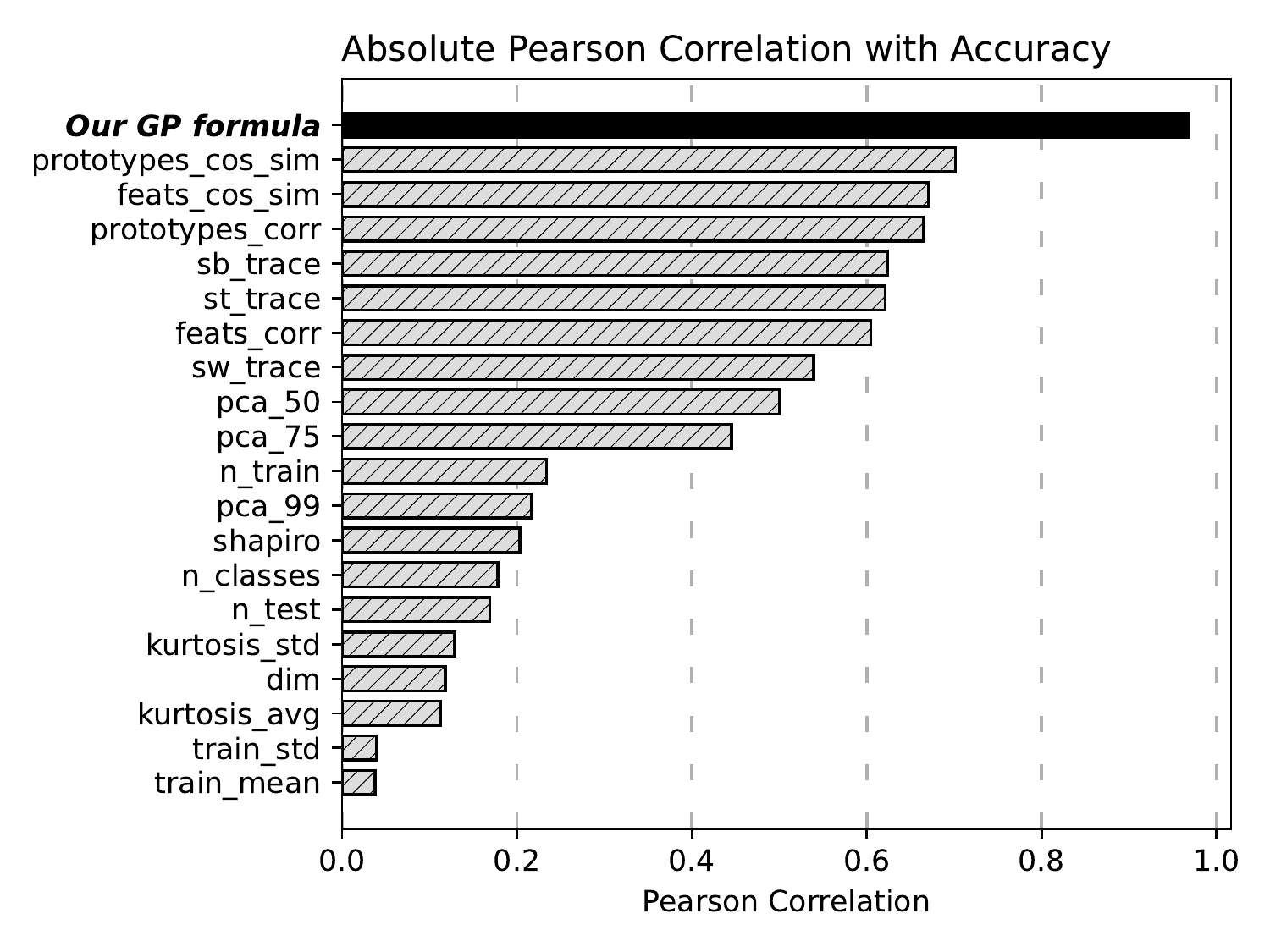}
    \caption{Absolute value of Pearson correlation between each dataset statistic and accuracy.} 
    \label{fig:abs_corr_temp} 
\end{figure}

~\\
\noindent
\textbf{Meta-Dataset Representation} To be able to  find the hidden relationship between a given dataset and the associated optimal accuracy, we need to describe each of those datasets by a feature vector $s$ in a 
shared representation 
space $\mathcal{S}$. 
Inspired by \cite{ho2002complexity, lorena2019complex, mansilla2004classifier, chamand2022heuristic} we selected various features $s_i$: the dimensionality of embeddings (\textit{dim}), 
the number of output classes (\textit{n\_classes}), 
the trace of the average matrix of all intra-class covariance matrices (\textit{sb\_trace}), 
the trace of the average of all inter-classes covariances matrices (\textit{sw\_trace}), 
the sum of the two previous traces (\textit{st\_trace}),  
the mean squared deviation (MSD) between the features' correlation matrix and the identity (\textit{feats\_corr}), 
 the mean cosine similarity between 
 each pair of dimensions  (\textit{feats\_cos\_sim}), 
the percentage of dimensions to be retained for a given explained variance 
of 50\%, 75\% and 99\% (\textit{pca\_XX\%}) to capture information about the dataset intrinsic dimension, the average of all embedding values (\textit{train\_mean}) and the standard deviation  (\textit{train\_std}), the average and the standard deviation of the  kurtosis computed on each dimension (\textit{kurtosis\_avg}, \textit{kurtosis\_std}), and the average Shapiro-Wilk value testing the normality of each dimension (\textit{shapiro}). 
The two variables (\textit{prototypes\_corr}, \textit{prototype\_cos\_sim}) refer respectively to information about the correlation and the cosine similarity between prototypes.
Here, the term prototype denotes the average embedding for each class.
Finally, we added number of samples in the training set (\textit{n\_train}) and the testing set (\textit{n\_test}).
The correlations of each statistic with accuracy are reported in Figure~\ref{fig:abs_corr_temp}.

\paragraph{Ground Truth Creation} 
Once we have extracted the embeddings from various datasets with feature extractors, we need to find the best reachable accuracy by a softmax classifier for each case.
To do so, we split each dataset of embeddings into training and testing sets and trained the model during 1000 epochs with a batch size of 2048. 
As pre-processing, all embeddings were only $\ell_2$-normalized.
The test sets are the usual ones for datasets with a specific split, such as CIFAR. 
We used a 66/33 split for few-shot datasets, such as THINGS, to ensure that the train/test split proportion left at least 10 images per class. 
The other ones were split with a ratio of 75/25. 
By tracking the accuracy on the test set, we can observe the best-reached accuracy $\alpha$ that we will consider as a good approximation of the best accuracy reachable $\alpha^*$. 
We used Adam optimizer.
Our meta-dataset $\mathcal{M} = \{(s_i, \alpha_i)\}_{i=1}^D$ corresponds to all the pairs of statistical representation $s_i \in \mathcal{S}$ of each dataset $d_i$ of the $D$ datasets and the observed  optimal accuracy $\alpha_i \in \mathcal{A}$. 
Those tuples are then our inputs and targets.

\begin{figure}
    \centering
    \includegraphics[width=.47\textwidth]{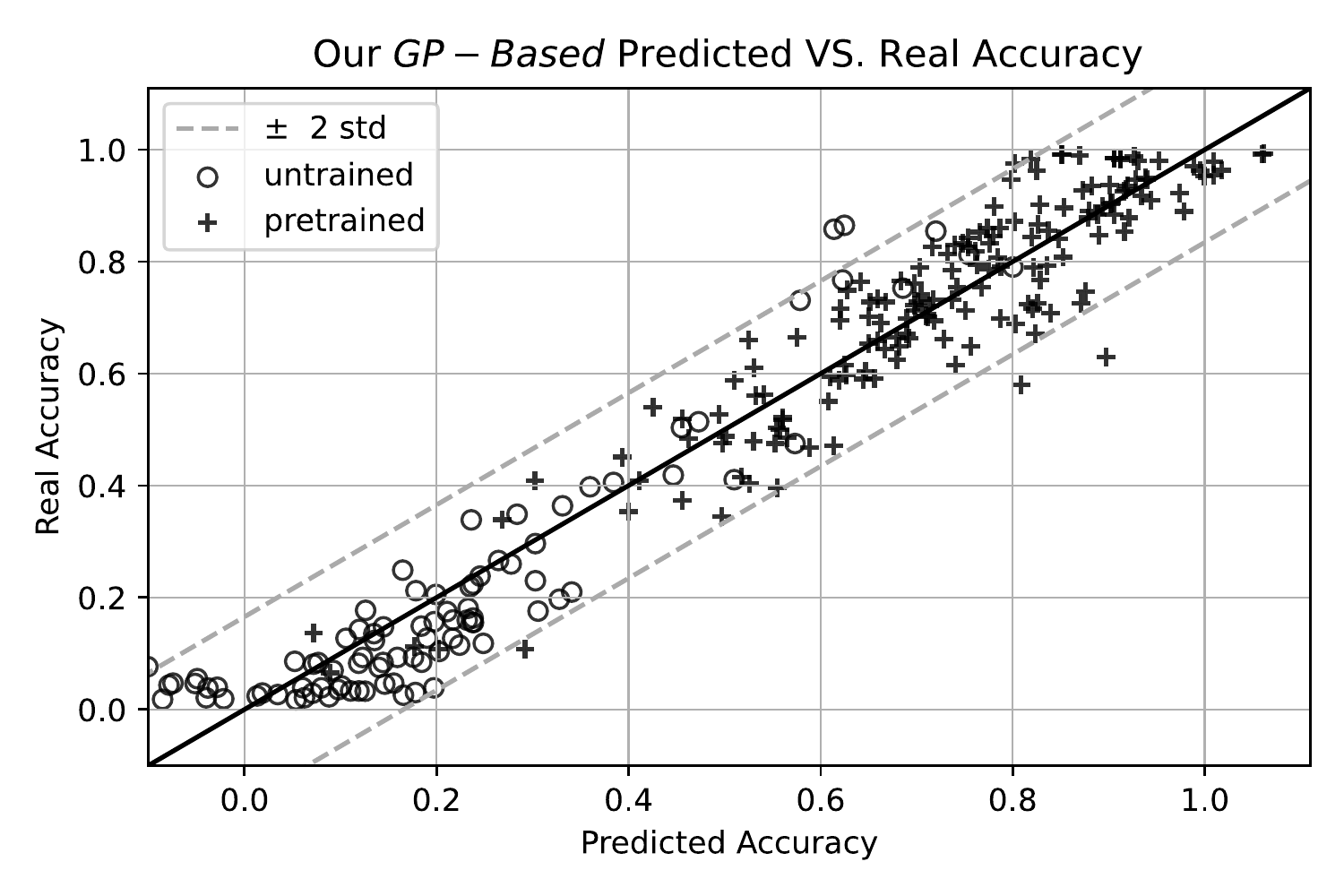}
    \caption{We can observe a strong linear relationship between our predicted accuracy compared to the real one.} 
    \label{fig:predicted_vs_real_accuracy}
\end{figure}

\paragraph{Symbolic Regression} 
Recovering hidden algebraic 
relationship between variables in order to describe a given phenomenon is the objective that symbolic regression (SR) seeks to optimize.
We search a prediction function $p : \R^S \rightarrow \R$ from our meta-dataset $\mathcal{M}$, with $S$ the number of statistical features representing each dataset $d_i$. 
As previously mentioned, different approaches have been developed for symbolic regression.
By benchmarking SR frameworks and ML models, it has been found that DSO \cite{petersen2021deep}, a deep learning-based approach, and \textit{gplearn}, a genetic programming (GP) framework, are two of the top-5 methods compared \cite{la2021contemporary}.
When trying with code provided by DSO \cite{petersen2021deep} on our task, solutions found under-performed the \textit{gplearn} with more complex formulas and longer training time.
Thus, we focus here on the \textit{gplearn} implementation\footnote{\url{https://gplearn.readthedocs.io/}} because of the compactness of the solutions found, speed of execution, and easiness of use. 
In GP-based symbolic regression, a population $\mathcal{P}$ of randomly generated mathematical expressions is "evolved" using evolutionary operations like \textit{selection}, \textit{crossover} and \textit{mutation} to improve a fitness function $\mathcal{F}$.
The individuals $p$ in the population $\mathcal{P}$ are represented as hierarchical compositions of primitive functions $\tau$ and terminals appropriate to the particular problem domain. 
Here, $\tau = \{ \log, e, \sqrt, +, -, \times, \div \}$ and the set of terminals corresponds to the statistics $s_i$ describing the dataset $d_i$. 
We evolved a population of 5000 individuals for 20 steps and tested 3 different fitness functions~: the first one corresponds to the $r^2$ between the predicted formula and the expected result.
This fitness function 
produced poor results both on training and testing sets.
The second one measures Pearson's correlation between the expected and predicted accuracies. 
While being easier to optimize than the first one, we found this one to be surprisingly inefficient since it tends to group the pretrained representations in a compact cluster, and the untrained 
ones in another one such that a line passes through the two centroids.
Indeed, the Pearson correlation between model accuracies and the variable specifying whether a pretrained or untrained model is used for embedding extraction is already at 0.77. 
To overcome this effect, we designed a simple fitness function such that both pretrained and untrained extracted embeddings independently have a linear correlation with accuracy.
For a given individual, here a GP predictor formula  $p(\cdot)$, we assess its fitness score $\mathcal{F}$: 
\begin{equation}
\begin{aligned}
    \mathcal{F} = \min\Big[ 
    & \big| \text{pearsonr}\Big(p(\mathcal{S}_{pretrained}),   \mathcal{A}_{pretrained}\Big) \big|, \\
    & \big| \text{pearsonr}\Big(p(\mathcal{S}_{untrained}),  \mathcal{A}_{untrained}\Big) \big| \Big]
\end{aligned}
\end{equation}
with $\mathcal{S}_{subset}, \mathcal{A}_{subset}$ corresponding respectively to the sets of statistical representations and target accuracies $\alpha$ of the given $subset \in \{pretrained, untrained\}$.
While this constraint does not enforce to have both pretained and untrained sets to be correlated with the same tendency, we can, however, experimentally observe the benefits 
on Figure~\ref{fig:predicted_vs_real_accuracy}, where pretrained and untrained networks are not separated in very distinct clusters but are distributed around a line.
We split our meta-dataset in a fixed 75/25-train/test fashion and repeated each experiment 1000$\times$.
Since $\mathcal{F}$ only seeks for correlation, a linear transformation of the output value is learned on the training set in order to predict the accuracy: $\hat{\alpha} = a \cdot p(\cdot) + b$.

\section{Results}


\begin{table}
\caption{Our formula has a better correlation and higher predictive power with only 5 variables, while the other models used the 19 variables (all $p$-value $<$ 0.01).} 
\centering
   \begin{tabular}{  l  c  c  }
     \toprule
     Method & Pearson$r$ & $r^2$  \\
     \midrule
     Linear Regression  & 0.9042 & 0.8011 \\ 
     Decision Tree  & 0.9472 & 0.8868 \\ 
     Random Forest (10 trees)  & 0.9643 & 0.9246 \\ 
     Our GP formula ($GPF$)  & \textbf{0.9671} &\textbf{0.9319} \\ \bottomrule
   \end{tabular}
 
  \label{tab:comparaison_with_other_regression_models}
\end{table}

\paragraph{Baselines} 
To evaluate the performance of our GP solution, we compare it with popular regression methods, including linear regression, decision tree regression, and random forest regression.
The same training/test split has been used for all those methods.
All variables are used simultaneously.
Performances on the test set are reported in Table.~\ref{tab:comparaison_with_other_regression_models}. 
With a substantial gap of $r^2$ score between the linear regressor and our formula, we can conclude that the task of predicting the accuracy requires a complex non-linear combination of only a few variables.
Furthermore, we compare with non-linear regressors such as decision trees and random forests.
We choose those because of their performances and the widespread belief suggesting those models are among the most interpretable ones. 
We used \textit{sklearn} implementations.
Our formula outperformed the decision tree and performed similarly to the random forest while being much more explainable.

\begin{figure}
    \centering
    \includegraphics[width=.47\textwidth]{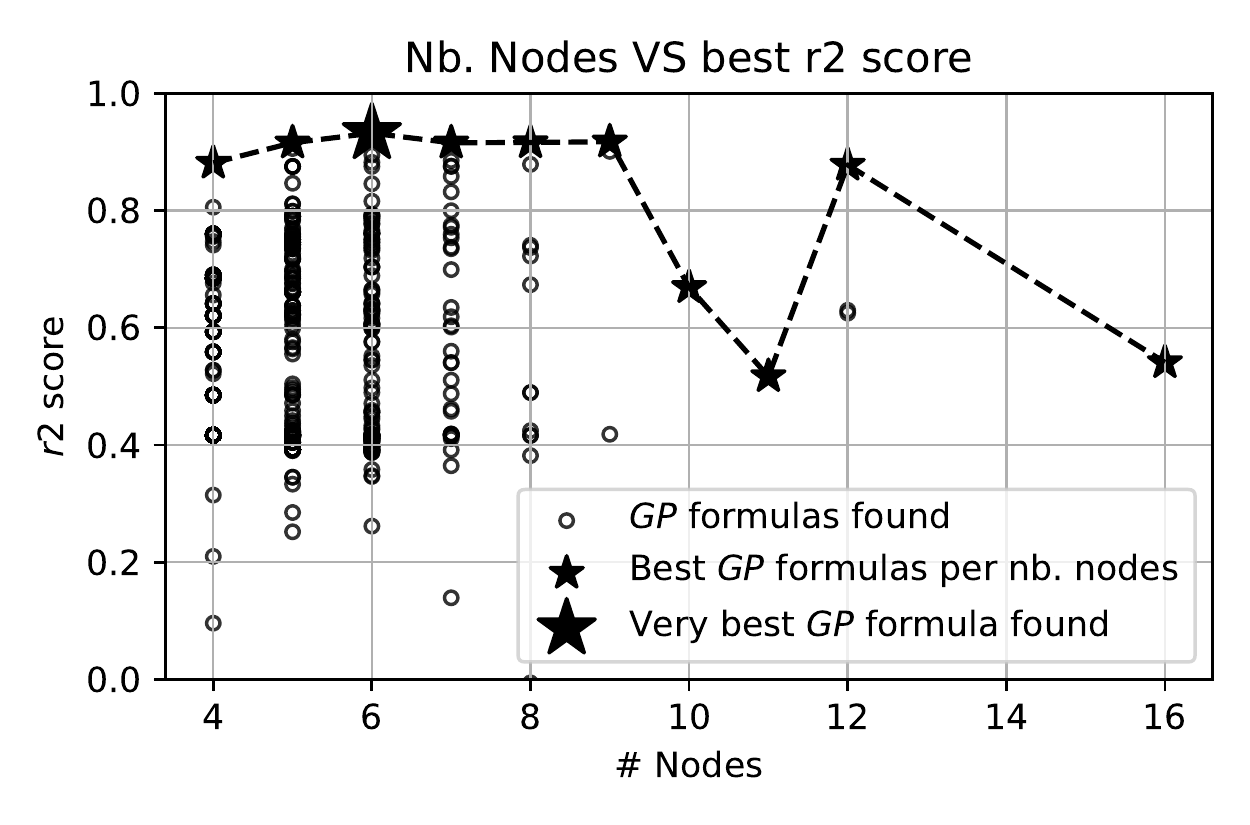}
    \caption{Performance versus the complexity of GP-Formulas.}
    \label{fig:nb_nodes_vs_r2}
\end{figure}

\begin{figure}
    \centering
    \includegraphics[width=.47\textwidth]{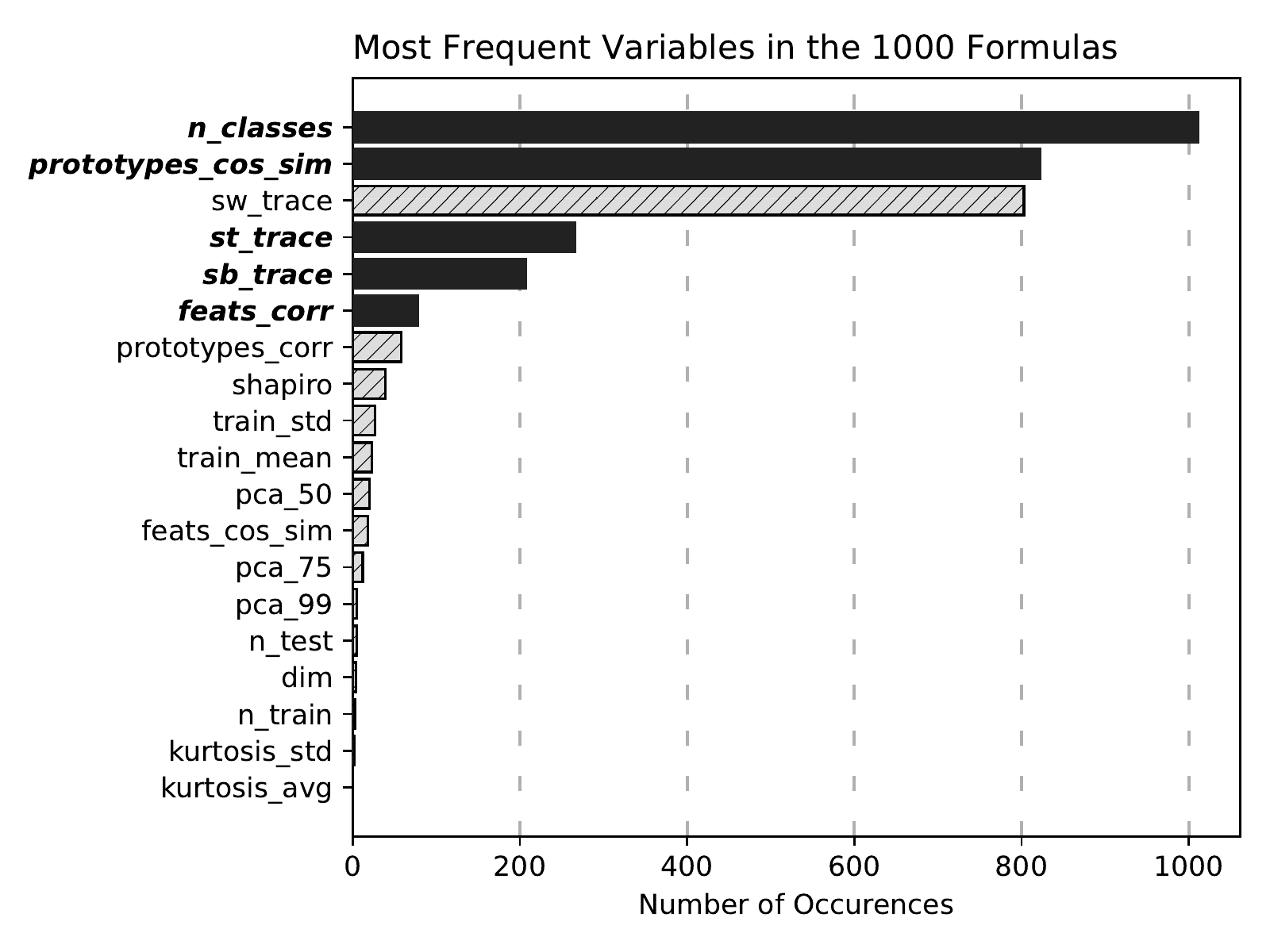}
    \caption{Frequency of the most used variables in the 1000 GP-formulas found. Dark bars indicate the variables present in the best GP formula.}
    \label{fig:most_used_variables}
\end{figure}

\paragraph{Symbolic Regression Formula} We ran our GP pipeline 1000 times on the same training set and serialized their respective solutions and scores for analysis.
The solution having the best test $r^2$ score was found $6\times$.
We compare on Figure~\ref{fig:nb_nodes_vs_r2} the test performances to the complexity of solutions found.
Our formula has a complexity of 6 nodes. 
We will refer to this \textit{\textbf{G}enetic \textbf{P}rogramming \textbf{F}ormulas} as: 
\begin{equation}
    \label{eq:gpftest}
        \resizebox{.9\hsize}{!}{${GPF} = \log\left(\frac{{sb\_trace} / {st\_trace}}{\sqrt{n\_classes \cdot feats\_corr \cdot prototypes\_cos\_sim}}\right)$}
\end{equation}
We can easily rewrite 
:  $GPF = SEP - COR$ with:
\begin{equation} 
    \begin{split}
    {SEP}   & = \log\left(\frac{sb\_trace}{st\_trace}\right) \\
    {COR} & = \frac{1}{2}\log\left(n\_classes \cdot feats\_corr  \cdot prototypes\_cos\_sim\right)
    \end{split}
\end{equation}
$SEP$ may correspond to a \textbf{\textsc{sep}}arability criterion while $COR$ may correspond to \textbf{\textsc{cor}}relation information. 
Section.~\ref{sec:discussion} delves deeper into each formula component. 
By ablating $GPF$ and considering each part independently, we found they were complementary.
Indeed, $SEP$ has only a Pearson's correlation of $0.65$ and $COR$ of $-0.87$ while the combination of the two parts reached $0.96$.
Finally, we found that other best-performing GP formulas have a similar structure and variables.
We report in Figure~\ref{fig:most_used_variables} how many times each statistic was used during the 1000 runs.

\begin{table}
    \caption{The slight variation in baseline precision when only using the five variables chosen by our genetic formula suggests that those five variables are the most important for all methods. (all $p$-value $<$ 0.01).} 
    \centering
       \begin{tabular}{lrr}
    \toprule
    Method &  Pearson$r$ &         $r^2$ \\
    \midrule
    Linear Regression         &  0.8796 &    0.7689 \\
    Decision Tree             &  0.9538 &    0.8937 \\
    Random Forest (10 trees)  &  0.9532 &    0.9057 \\
    Our GP formula ($GPF$)  & \textbf{0.9671} &\textbf{0.9319} \\
    \bottomrule
    \end{tabular}
     
      \label{tab:comparison_only_5_vars}
\end{table}

\begin{table}
\caption{We evaluate the performance of our formula after replacing each variable with its mean value. All $p$-values $<$ 0.01, except for Sb\_trace with a $p$-value of $0.6882$. Bold scores correspond to the worst one making the variable the most important one.}
    \centering
        \begin{tabular}{lrr}
        \toprule
        Ablated Variable &   Pearson$r$ &        $r^2~~~$ \\
        \midrule
        \textit{Sb\_trace}            &  \textbf{$-$0.0503} & \textbf{$-$2.1806} \\
        \textit{n\_classes}           &  ~0.7918 & $~$0.4341 \\
        \textit{St\_trace}            &  ~0.8028 & $-$1.2761 \\
        \textit{feats\_corr}          &  ~0.8530 & ~0.5818 \\
        \textit{prototypes\_cos\_sim} &  0.9420 & 0.8764 \\
        \midrule
        No variable ablated       &  \underline{0.9671}  &  \underline{0.9319}  \\
        \bottomrule
        \end{tabular}
    
    \label{tab:ablation_variables}
\end{table}

\paragraph{Ablation} 
As a first ablation test, we propose demonstrating how the variables selected by our $GPF$ influence the results for different baselines. 
Table.~\ref{tab:comparison_only_5_vars} shows a minor decrease in performance of those methods when compared to the original baseline from Table.~\ref{tab:comparaison_with_other_regression_models}.
For example, Pearson's correlation of the random forest falls from $0.9643$ to $0.9532$, while the decision tree rises from $0.9472$ to $0.9538$. 
On the other hand, the linear regression model needs to suffer from this feature selection step by dropping from $0.9042$ to $0.8796$. 
All $p$-values $<$ 0.01 suggests that selected variables could be sufficient to correlate with accuracy. 
However, a non-linear transformation of those variables is still required.

One can note that the log properties can still be applied to our $GPF$ to reduce our formula as a linear combination of our selected variables. 
With coefficients being $[1, -1, -\frac{1}{2}, -\frac{1}{2}, -\frac{1}{2}]$, our $GPF$ could thus be written as: 
\begin{equation}
    \begin{split}
        \texttt{GPF =} 
            & \texttt{~log(Sb\_trace) - log(St\_trace) } \\
            & \texttt{ - 0.5 log(n\_classes)} \\
            & \texttt{ - 0.5 log(feats\_corr)} \\
            & \texttt{ - 0.5 log(prototypes\_corr)}
    \end{split}
    \label{eq:our_GPF_seen_as_combination}
\end{equation}
Given that no prior on the structure of the $GPF$ was imposed during the search phase, this discovery is particularly interesting. 
With this finding arise two closely related questions: 
1) does learning those five coefficients improve performance?
2) what would be the performances if all variables were used? 
To answer the first one, we learned a linear regression model on the five statistics after passing them to the log. 
Doing so increased the performance of the linear model without the log transform from a Pearson's coefficient of 0.9042 in Table.~\ref{tab:comparaison_with_other_regression_models} to 0.9607 after log transforming inputs as reported in Table.~\ref{tab:comparison_to_related_work_test}. 
While being more efficient on the train set, the linear model performed worse on the test set than our $GPF$. 
However, when comparing the learned weights, we found that signs and magnitudes were highly similar to weights or our $GPF$ with a cosine similarity of $0.9923$. 
By not requiring any re-weighting of our five variables, our formula in its original form (Equation.~\ref{eq:gpftest}) is thus more interesting. 
On the other hand, we learned a linear model on the log-transformed statistics using the same procedure. 
Due to negative values in the original ones, only 17 of 19 are kept. 
At the expense of being significantly less explainable, the model's scores are reported in Table.~\ref{tab:comparaison_with_other_regression_models} outperformed our $GPF$ ones. 
Ones can note that the relative difference of reported correlations between the models with 17 variables and the $GPF$ with 5 is smaller than the difference of the $GPF$ with the 5 variables and the $GPF$ with the 4 most important variables reported in Table.~\ref{tab:ablation_variables}.

Finally, our last ablation study seeks to determine which components of our formula are the most important ones. 
We can determine how much each variable influences scores by freezing each variable and replacing it with its mean value. 
The more significant the drop, the greater the variable's significance. 
Table.~\ref{tab:ablation_variables} allows us to see that freezing each variable results in a decrease in score. 
All $p$-values are significant ($<$ 0.01), with the exception of \textit{Sb\_trace}. 
Indeed, freezing it removes all correlation between our $GPF$ and the expected accuracy with a $p$-value of $0.6882$. 
On the other hand, while having a minimal positive impact on correlating $GPF$ with accuracy, \textit{prototypes\_cos\_sim} appears to be still important to have a good $r^2$ prediction score.

\begin{table}
\caption{We compare our solution to different methods. Our $GPF$ offers the best compromise between accuracy and simplicity without having any prior on the structure of the solution. Our extension of the $GPF$ using 17 variables offers the best Pearsons's correlation and $r^2$ score. Here \textit{log} stands for log-transformed variables and \textit{org} for the original ones.}
\centering
\begin{tabular}{lrcr}
\toprule
             Method   &  Nb. Var      & Pearson$r$               & $r^2$~~          \\ \midrule
GA unweighted sum & 19 org &  0.7763  &  0.5744 \\
GA unweighted sum & 17 log &  \textit{0.9621}  &  \textit{0.9254} \\
\midrule
Cubist Rules  & 19 org          & 0.9666                   & 0.9343    \\  
Cubist Rules & ~5 log          & 0.9642                   & 0.9276            \\ 
Cubist Rules & 17 log          & \textit{0.9772}                   &    \textit{0.9525}         \\ 
\midrule
Our Linear Regression & 5 log & 0.9607                  & 0.9206      \\
Our Linear Regression & 17 log & \textbf{0.9795}                  & \textbf{0.9586}      \\
Our GP formula    & 5 org    & \underline{\textit{0.9671}}          & \underline{\textit{0.9319}}  \\ \bottomrule
\end{tabular}
\label{tab:comparison_to_related_work_test}
\end{table}

\paragraph{Comparison to Related Work} 
As previously mentioned, \cite{Collins2018EvolutionaryDM, bensusan2001estimating} used neither similar datasets nor statistics for describing selected datasets making their work hard to compare them and with them. 
However, we propose to compare our 
solutions found by applying their pipeline on our meta-dataset. 
Results are reported in Table.~\ref{tab:comparison_to_related_work_test}.

To find an unweighted sum of a few variables, we used a similar \textit{genetic algorithm}\footnote{\url{https://github.com/rmsolgi/geneticalgorithm}} (GA) with the Pearson's correlation between predicted and real score as fitness function, such as in  \cite{Collins2018EvolutionaryDM}. 
By setting variable type being integer and bounded between $[-1, 1]$, the 3 possibles values are $\{-1, 0, 1\}$. 
With an initial population of 5000 and 300 iterations, we found results to be stable. 
We used the exact same train/test split as us. 
To compute the $r^2$ score, we employed the same procedure of linearly re-calibrating the formula with $(a, b)$ learned on the training set. 
As expected, the results on the 19 variables are significantly worse than the weighted summation of our baseline linear regression model. 
Thus, we tested the same pipeline after keeping and log transforming 17 variables (due to 2 out of the 19 variables having negative values). 
Consistently with the discovery of the ablation study suggesting only to log transforming variables as pre-processing, results increased. 
The best solution found with the genetic algorithm (GA) is reported in Equation.~\ref{eq:best_ga_found}:  
\begin{equation}
    \begin{split}
    \texttt{GA =} & \texttt{~log(Sb\_trace) + log(shapiro)} \\
    & \texttt{ + log(dim) - log(feats\_corr)} \\
    & \texttt{ - log(Sw\_trace)  - log(kurtosis\_avg) } \\
    & \texttt{ - log(prototypes\_corr)}
    \end{split}
    \label{eq:best_ga_found}
\end{equation}
Solution found used 7 variables while our $GPF$ used only 5. 
With only 3 variables shared with our $GPF$, we find it difficult to understand the interaction with the selected variables.

\begin{figure}
    \centering
\resizebox{.9\hsize}{!}{\fbox{\parbox{.5\textwidth}{
    \texttt{
if \\
{\color{white} .} percentage\_dims\_exp\_var\_99 > 0.9316406 \\
{\color{white} .} feats\_corr <= 0.05039461 \\
{\color{white} .} kurtosis\_std <= 9.284021 \\
{\color{white} .} n\_test > 1880 \\
{\color{white} .} n\_test <= 4384 \\
{\color{white} .}then \\
{\color{white} .} outcome = 5.4992777 \\
{\color{white} .} ~~~- 0.031372 kurtosis\_std \\
{\color{white} .} ~~~- 5.02 percentage\_dims\_exp\_var\_99 \\
{\color{white} .} ~~~+ 1.31 feats\_corr \\
{\color{white} .} ~~~+ 0.83 Sb\_trace \\
{\color{white} .} ~~~- 0.19 prototypes\_cos\_sim \\
{\color{white} .} ~~~+ 0.085 prototypes\_corr \\
{\color{white} .} ~~~+ 8e-06 n\_test + 1.6 train\_mean \\
{\color{white} .} ~~~- 3.2e-05 n\_classes \\
{\color{white} .} ~~~- 1.8 train\_std
    } 
}}
}
    \vspace{1mm}
    \caption{Example of one of the ten rules output by \textit{Cubist} when learning on the 19 variables.} 
    \label{fig:example_cubist}
\end{figure}

To compare our solution to the pipeline proposed by \cite{bensusan2001estimating}, we used the R package implementing Cubist, the software the authors used to find an interpretative set of rules. 
As reported in Table.~\ref{tab:comparison_to_related_work_test} we experimented with three set of input variables. The first one corresponds to our 19 original variables without any transformation. 
While having scores comparable to our $GPF$ using 5 variables, we can note that the rules are highly complex. 
Indeed, it produced 10 rules, using many coefficients, which are hard to read. 
One example of rule is reported in Figure.~\ref{fig:example_cubist}. 
In a second experiment, we used only the top 5 variables selected by our $GPF$ after log transforming them. 
It helped the Cubist system in outputting comparable results by using only two rules. 
Each rule predicts the accuracy as a linear combination of all 5 log variables. 
We compare the coefficients of those rules with those of our original $GPF$ (as in Equation.~\ref{eq:our_GPF_seen_as_combination}. 
Interestingly, they have a cosine similarity of $0.9467$ and $0.9791$ with our $GPF$ ones. 
With only 5 variables, our $GPF$ is simpler and performs better. 
Finally, we found that our log pre-processing also benefited Cubist. 
When giving the 17 log-transformed variables as input, Cubist proposed a solution based on 6 rules while being significantly more efficient than the 10 rules outputted from the 19 original variables. 
However, our extension of the $GPF$ to the linear combination of the 17 log-variable still performs better while being much more straightforward than the Cubist's solutions.

Using Ockham's principle, those findings are evidence making our $GPF$ a better choice. 
Furthermore, our $GPF$ can be easier to explain because of its conciseness. 
We propose to discuss our $GPF$ in Section.~\ref{sec:discussion}.

\begin{table}
\caption{We examine our formula's transferability of baseline regressors with five variables. The best reachable $GPF$ is indicated with the \textbf{*} symbol. All $p$-values $<$ 0.01.}
\centering
    \begin{tabular}{lcr}
        \toprule
        Method &  Pearson$r$ &        $r^2~~$ \\
        \midrule 
        Linear Regression      &  0.6191 &  0.3052 \\
        Decision Tree &  0.7944 &  0.1928 \\
        Random Forest (10 trees) &  0.7231 & -0.0722 \\
        Our GP formula ($GPF$) & \textbf{0.8618} & \textbf{0.4565} \\
        \midrule
        Our GP formula ($GPF$)~\textbf{*} & \textbf{0.8618} & \textbf{0.7428} \\
        \bottomrule
    \end{tabular}
    \label{tab:transferability_of_regressor_models_to_text}
\end{table}

\paragraph{Generalization}
Our method is applicable to any dataset describable with a set of a few statistics. 
Because we obtained our $GPF$ using statistics from embedding datasets extracted only from vision datasets and feature extractors, the generalization of our $GPF$ discovered on those datasets to other domains, such as text, can be questioned. 
Thus, we used 7 text datasets, and 4 pretrained text features extractors from the \textit{sentence-transformers} package\footnote{\url{https://www.sbert.net}} to test our formula's ability to transfer to new modalities. 
Combining all those datasets and feature extractors, we applied the same process to extract dataset statistics and accuracies, yielding 28 points for our analysis. 
Table.~\ref{tab:transferability_of_regressor_models_to_text} compares how our formula transfers to this new set of points with classical regressors. 
All reported correlations have a statistically significant $p$-value ($<$ 0.01). 
We can observe a significant drop in Pearson's correlation and $r^2$ scores for all methods. 
However, our $GPG$ still outperforms other methods with a strong Pearson's correlation of 0.8618. 
Two $r2$ scores for our $GPF$ were reported; the first one corresponds to our formula linearly transformed with the coefficient $a, b$ learned on the training set of the vision dataset. 
The second one, obtained by linearly translating our $GPF$, refers to the best possible $r^2$ score on the text meta-dataset. 
To find the \textit{oracle} coefficients, we evaluated the formula after learning the parameters $a, b$ on the text meta-dataset (here train$=$test). 
The oracle gives us the best score reachable on this meta-dataset of 28 points. 
With $(a, b) = (0.2417, 1.0327)$ for the vision meta-dataset and $(a, b) = (0.2508, 0.9121)$ for the text meta-dataset, we can see that the parameters from the text and the vision meta-dataset are similar. 
However, the $r2$ score appears extremely sensitive, perhaps due to the small number of points. 
This score drop can be explained by the discrepancy between the text and vision meta-datasets. 
For example, while datasets with more than 200 classes are common in vision, text classification tasks typically have a much lower number of classes, such as 2 for sentiment analysis or 20 for topics modeling. 
We measured this discrepancy by performing  Student's $t$-test on each of the five selected variables. 
We found three of the five variables have a $p$-value $<$ 0.01 (\textit{Sb\_trace, n\_classes, prototypes\_cos\_sim}), giving evidence against the null hypothesis of equal population means. 
While not perfect, our results appear promising. However, they would benefit from incorporating more meta-datasets from other domains, such as audio, video, graph-based, or tabular data classification datasets.

\section{Discussion}
\label{sec:discussion}

As seen previously, $GPF$ can be written as a summation of two components.
With a closer look, one can observe that the first element $SEP$ is close to the Fisher's criterion used in the \textit{Linear Discriminant Analysis} (LDA) \cite{fisher1936use} where the objective is to find a linear projection that maximizes the ratio of between-class variance and the within-class variance.
Thus, $SEP$ corresponds to a separability measure of classes.
Interestingly, this criterion has been used successfully as a loss function in deep learning \cite{DorferKW15deeplda, ghojogh2020fisher}.
The choice of an LDA-based loss function remains marginal in deep learning, the cross-entropy (CE) being a more popular choice.
However, strong similarities between the LDA and the CE allow us to swap this first separability measure for the latter.
Indeed, \cite{wan2018rethinking} noticed that one of the most widely studied technical routes to overcome certain deficiencies of the softmax in the cross-entropy-based loss is to encourage stronger intra-class compactness and larger inter-class separability, analogously to Fisher's criterion.

The second part, $COR$, is negatively correlated to the accuracy. 
This is easily understandable by looking at each variable composing this part of the formula.
The first is the number of classes ($n\_classes$). 
Indeed, when a machine learning model is trained on a dataset, it is natural to expect that scores will decrease as the number of classes grows. 
\cite{gupta2014training} discuss how, as the number of classes in a dataset grows, it gets harder to distinguish between them, making the dataset increasingly challenging to classify. 
This intuition may be empirically verified on datasets with different class granularities.
For example, \cite{chang2021your} observed a drop in accuracy from 0.97 to 0.82 on the CUB200 dataset \cite{WelinderEtal2010}
when changing the number of classes from a coarse level (13) to a fine-grained one (200).
The two other variables ($feats\_corr$, $prototype\_cos\_sim$) correspond to  orthogonality and decorrelation information. 
By looking at the literature, we can easily explain the importance of both decorrelation terms. 
In defense of the weights decorrelation term (\textit{prototypes\_cos\_sim}), \cite{bansal2018ortho} found on several state-of-the-art CNN that they could achieve better accuracy, more stable training, and smoother convergence by using orthogonal regularization of weights. 
Previous works on features decorrelation heavily justify the presence of our features decorrelation variable (\textit{feats\_corr}) \cite{bardes2022vicreg, ermolov2021whitening, hua2021feature, kessy2018optimal, LeCun2012, wan2018rethinking,  zhang2021zero}. 
Indeed, \cite{LeCun2012} found that correlated input variables usually lead the eigenvectors of the Hessian to be rotated away from the coordinate axes leading to slower convergence. 
Thus, several propositions were developed to better decorrelate variables such as PCA or ZCA \cite{kessy2018optimal}. 
More recently, decorrelation played an essential role in the performance increase of self-supervised methods \cite{bardes2022vicreg, ermolov2021whitening, hua2021feature, zhang2021zero}. 
For example, \cite{ermolov2021whitening} recently introduced a whitening step in their self-supervised loss, and \cite{bardes2022vicreg} included a decorrelation part in their loss. 
They argue that this term decorrelates the variables and prevents collapse.

\section{Conclusion}

In this paper, we showed that a simple pipeline could help us to extract theoretical intuitions from experimentation.
To do so, we conducted experiments on a meta-dataset of more than 260 datasets of embeddings extracted from the combination of a wide range of datasets and feature extractors.
To solve the problem of expressing such disparate datasets, we proposed combining them into a single space by creating a representation using a set of general statistics that can be computed on any dataset.
As a result, our work applies to computer vision and all other areas of machine learning. 
Finally, an heuristic able of predicting the accuracy of a linear classifier was discovered automatically, with a Pearson's correlation of $0.96$ and an $r^2$ of $0.93$.
Interestingly, other systems with similar performances tend to confirm our $GPF$ by having highly correlated weights. 
Furthermore, this formula is highly explainable and is consistent with decades of research. 
This successful example of AI-assisted research encourages us to use it in other areas, such as predicting and understanding hyperparameters  (regularization, temperature, tree depth, etc.).

\section{Acknowledgment}
    
This work was partially financed by \textit{Smiths Detection}. \\
Authors would like to thank all peoples involved in the proof-reading and contributed to substantially improving this document. 
Listed in alphabetical order: 
Thibault ALEXANDRE, Ihab BENDIDI, Mohamed CHELALI, Philippe JOLY, Celia KHERFALLAH, Camille KURTZ, 
Amine MARZOUKI, Julien PINQUIER, Guillaume SERIEYS.


\printbibliography

\end{document}